\documentclass{article}

\usepackage[nonatbib,preprint]{neurips_2025}

\usepackage[utf8]{inputenc} \usepackage[T1]{fontenc}    \usepackage{hyperref}       \usepackage{url}            \usepackage{booktabs}       \usepackage{amsfonts}       \usepackage{nicefrac}       \usepackage{microtype}      \usepackage{xcolor}         

\usepackage{amsthm}         \usepackage{amsmath,amssymb,enumerate,color}
\usepackage{multirow}       \usepackage{threeparttable} \usepackage{graphicx}
\usepackage{makecell}

\usepackage{subfigure}
\usepackage{enumitem}
\usepackage{adjustbox}

\title{Recurrent Neural Operators: Stable Long-Term PDE Prediction}

\author{
  Zaijun Ye\\
  Shanghai Normal University\\
  Shanghai, China\\
  \texttt{yezaijun@outlook.com} \\
  \And
  Chen-Song Zhang\\
  Chinese Academy of Sciences\\
  University of Chinese Academy of Sciences\\
  Beijing, China\\
  \texttt{zhangcs@lsec.cc.ac.cn} \\
  \And
  Wansheng Wang\thanks{Corresponding author.}\\
  Shanghai Normal University\\
  Shanghai, China\\
  \texttt{w.s.wang@163.com} \\
}

\newtheorem{theorem}{Theorem}
\newtheorem{lemma}{Lemma}
\newtheorem{proposition}{Proposition}

\begin{document}

\maketitle

\begin{abstract}
  Neural operators have emerged as powerful tools for learning solution operators of partial differential equations. However, in time-dependent problems, standard training strategies such as teacher forcing introduce a mismatch between training and inference, leading to compounding errors in long-term autoregressive predictions. To address this issue, we propose Recurrent Neural Operators (RNOs)—a novel framework that integrates recurrent training into neural operator architectures. Instead of conditioning each training step on ground-truth inputs, RNOs recursively apply the operator to their own predictions over a temporal window, effectively simulating inference-time dynamics during training. This alignment mitigates exposure bias and enhances robustness to error accumulation. Theoretically, we show that recurrent training can reduce the worst-case exponential error growth typical of teacher forcing to linear growth. Empirically, we demonstrate that recurrently trained Multigrid Neural Operators significantly outperform their teacher-forced counterparts in long-term accuracy and stability on standard benchmarks. Our results underscore the importance of aligning training with inference dynamics for robust temporal generalization in neural operator learning.
\end{abstract}

\section{Introduction}

Evolution equations constitute the foundation for modeling dynamic systems across diverse scientific and engineering disciplines, including physics, biology, finance, and fluid dynamics. These equations describe how systems evolve over time. A general form of such an equation can be expressed as:
\begin{equation}\label{eq:base_problem}
    \left\{  
    \begin{aligned}
        \frac{\partial u}{\partial t} &= \mathcal{D}(u, f_1) , && \text{in} &&[0,+\infty) \times \Omega, \\  
        u(0, x) &= f_2(x), && \text{in} &&\;\Omega, \\ 
        \mathcal{B}u(t, x) &= f_3(t, x), && \text{on} &&[0,+\infty) \times \partial \Omega, \\ 
    \end{aligned}  
    \right.  
\end{equation}  
where \( u \in \mathbb{U} \) denotes the state variable (solution), \( \Omega \) is the spatial domain, and \( t \) represents time. The function \( f_1 \in \mathbb{F}_1 \) represents parameters or forcing terms, \( f_2 \in \mathbb{F}_2 \) specifies the initial state at \( t=0 \), and \( f_3 \in \mathbb{F}_3 \) defines the boundary conditions on the boundary \( \partial \Omega \). The operator \( \mathcal{B}: \mathbb{U} \mapsto \mathbb{F}_3 \) is a boundary operator. The operator \( \mathcal{D} : \mathbb{U} \times \mathbb{F}_1 \mapsto \mathbb{D} \) is potentially a nonlinear differential operator, where \( \mathbb{U}, \mathbb{F}_1, \mathbb{F}_2, \mathbb{F}_3, \) and \( \mathbb{D} \) are suitable Banach spaces. For notational simplicity, we bundle the input functions as $f := (f_1, f_2, f_3) \in \mathbb{F} := \mathbb{F}_1 \times \mathbb{F}_2 \times \mathbb{F}_3$.

Obtaining analytical solutions for Equation \eqref{eq:base_problem} is often intractable. While traditional numerical methods (e.g., FDM, FEM) provide robust simulation frameworks through spatio-temporal discretization, they can face challenges with computational expense, mesh complexity, stability constraints, and high dimensionality. Early deep learning methods for PDEs, such as PINNs \cite{raissiPhysicsinformedNeuralNetworks2019}, DGM \cite{sirignanoDGMDeepLearning2018}, Deep BSDE solvers \cite{eDeepLearningBasedNumerical2017}, as well as approaches like those in  \cite{raissiForwardBackwardStochastic2023,eDeepRitzMethod2018, hureDeepBackwardSchemes2020, yeFBSJNNTheoreticallyInterpretable2024} primarily focus on approximating the solution for a single instance (fixed \(f\)).

To address the need for solving entire families of PDEs parameterized by input functions \( f \), the paradigm of operator learning has gained significant traction \cite{kovachkiNeuralOperatorLearning2023}. Operator learning methods aim to approximate the solution operator from the input function space \( \mathbb{F} \) to the solution space \( \mathbb{U} \). DeepONet derivatives trace to the foundational neural operator framework \cite{luLearningNonlinearOperators2021}, later extended to ensure reliable extrapolation via physics-informed constraints \cite{zhuReliableExtrapolationDeep2023}. Spectral-aware architectures originate from the Fourier Neural Operator (FNO) \cite{liFourierNeuralOperator2021}, with extensions including amortized parameterizations \cite{xiaoAmortizedFourierNeural2024}, factorized forms \cite{tranFactorizedFourierNeural2023}, and adaptations for hyperbolic conservation laws \cite{kimApproximatingNumericalFluxes2025}. Replacing the Fourier transform with the Laplace transform leads to the Laplace Neural Operator (LNO) \cite{caoLNOLaplaceNeural2024}. Mesh-aware architectures, grounded in classical numerical methods, include the Multigrid Neural Operator (MgNO) \cite{heMgNOEfficientParameterization2024}, and hierarchical matrix-inspired approach \cite{liuMitigatingSpectralBias2024}. These approaches offer the potential for substantial computational speed-ups at inference time: once the operator is trained, solutions for new instances \( f \) can be obtained rapidly via a single forward pass through the network.

Applying operator learning to evolution problem \eqref{eq:base_problem} is particularly compelling. Many real-world systems, such as weather forecasting \cite{mccabeStabilityAutoregressiveNeural2023, pathakFourCastNetGlobalDatadriven2023} or reservoir simulation \cite{liuMgNOMethodMultiphase2025}, are governed by complex dynamics where either the governing equations are imperfectly known or historical data is abundant. Operator learning enables learning the solution operator, or even the underlying dynamics operator \( \mathcal{D} \) itself, potentially end-to-end from observational or simulation data. This bypasses explicit discretization steps and may mitigate errors arising from imperfect model assumptions.

However, a critical challenge emerges when employing operator learning for long-term prediction of evolution equations. Typically, models are trained to predict the solution \( u(t+\Delta t) \) based on the state \( u(t) \) and parameters \( f \), using ground-truth data available only within a finite time horizon \( [0, T] \). This training often relies on minimizing a one-step prediction loss, effectively implementing a strategy analogous to "teacher forcing" in sequence modeling, where the true state \( u(t) \) is provided as input at each training step. During inference, particularly for predicting beyond the training horizon, i.e., extrapolation in time, \( t > T \), the model must operate autoregressively: the prediction at time \( t \), denoted \( \hat{u}(t) \), serves as the input for predicting \( \hat{u}(t+\Delta t) \). This mismatch between the training regime (using ground truth) and the inference regime (using model predictions) creates a distribution shift. Consequently, prediction errors inevitably accumulate over successive time steps, potentially leading to instability and severely degrading the accuracy of long-term forecasts \cite{mccabeStabilityAutoregressiveNeural2023,wuCOASTIntelligentTimeAdaptive2025}.

While several recent studies have attempted to mitigate this error accumulation through techniques like post-hoc refinement, fine-tuning, or adaptive time-stepping strategies \cite{xuTransferLearningEnhanced2023, lippePDERefinerAchievingAccurate2023, wuCOASTIntelligentTimeAdaptive2025, michalowskaNeuralOperatorLearning2024}, these often introduce additional computational overhead or may not fundamentally resolve the train-test mismatch inherent in the standard teacher forcing paradigm.

In this paper, we target the challenge of learning robust dynamics operators for evolution equations, specifically aiming for accurate and stable long-term extrapolation ($t > T$) from data observed over a finite interval ($[0, T]$). We hypothesize that mitigating the mismatch between standard training protocols and the demands of autoregressive inference is key. To this end, we propose and systematically evaluate \textbf{Recurrent Neural Operators (RNOs)}, trained using a \textbf{recurrent structure}. This structure involves recursively utilizing the model's own predictions as inputs for subsequent steps within the training horizon $[0, T]$, directly reflecting the conditions encountered during inference. Our central thesis is that this recurrent training strategy significantly enhances the stability and accuracy of long-time extrapolation compared to conventional teacher forcing methods.

The remaining sections are organized as follows. Section~\ref{sec:related_work} reviews prior work on operator learning for evolution equations and the challenge of long-term prediction. Section~\ref{sec:methodology} presents the neural operator architectures, our proposed recurrent training framework, and the theoretical comparison of error propagation under recurrent versus teacher forcing strategies. Section~\ref{sec:experiments} outlines the main experimental results, evaluates rollout stability over long horizons, and includes an ablation study analyzing the impact of key components. Section~\ref{sec:conclusions} summarizes our findings and discusses potential limitations. Additional theoretical details and problem definitions are provided in the appendices. 
\section{Related Work}
\label{sec:related_work}
Our research focuses on advancing operator learning for time-dependent problems, particularly evolution equations, with a central emphasis on addressing the critical issue of error accumulation in long-term autoregressive prediction.

\subsection{Operator Learning for Evolution Equations}
Originally developed for parametric PDEs, operator learning aims to approximate mappings between infinite-dimensional function spaces. Foundational studies have established the universal approximation theorem for neural operators \cite{luLearningNonlinearOperators2021, kovachkiUniversalApproximationError2021}. Subsequent research has refined these foundational methods by incorporating physical priors \cite{liGeometryInformedNeuralOperator2023}, devising improved training strategies \cite{kimDPMNovelTraining2021}, improving computational efficiency and scalability to higher resolutions \cite{tranFactorizedFourierNeural2023, xiaoAmortizedFourierNeural2024, heMgNOEfficientParameterization2024}, deriving theoretical error bounds \cite{kovachkiUniversalApproximationError2021, rezaeiFiniteOperatorLearning2024}, and extending generalization capabilities \cite{liFourierNeuralOperator2023, zhuReliableExtrapolationDeep2023}.

When applied to evolution equations of the form \eqref{eq:base_problem}, operator learning approaches can generally be classified into three main categories:
\begin{itemize}[left=1.5em]
    \item \textbf{Spatio-temporal Unified Approach}: Treats time as an additional spatial coordinate, learning a direct map from inputs (initial conditions, parameters) to the solution field across the entire spatio-temporal domain (e.g., employing the FNO-3D mentioned in \cite{liFourierNeuralOperator2021}).
    \item \textbf{Time-as-Parameter Approach}: Includes time \(t\) as an explicit input coordinate to the neural operator, enabling the querying of the solution state at arbitrary time instances \cite{liuMgNOMethodMultiphase2025}.
    \item \textbf{Time-Stepping (or Autoregressive) Approach}: Learns an operator that propagates the solution forward by one or more discrete time steps, effectively approximating the underlying evolution dynamics \cite{liFourierNeuralOperator2021, heMgNOEfficientParameterization2024}.
\end{itemize}
Although the unified and time-as-parameter methods perform well within the training time horizon \( [0, T] \), the time-stepping approach better reflects the sequential and causal nature of temporal dynamics. This characteristic makes it potentially better suited for long-term prediction and extrapolation beyond the training window (\( t > T \)), which constitutes the primary focus of our work.

\subsection{Challenges in Long-Term Prediction and Extrapolation}
\label{subsec:related_challenges}
A significant obstacle in deploying time-stepping operator learning models, particularly for extrapolation beyond the training horizon \( T \), is the accumulation of errors during autoregressive prediction \cite{mccabeStabilityAutoregressiveNeural2023}. As highlighted in the Introduction, this stems largely from the mismatch between teacher-forcing during training and autoregressive inference at test time. Several strategies have been proposed to mitigate this phenomenon. Some efforts concentrate on enhancing the accuracy of the single-step predictor. For example, PDE-Refiner employs a diffusion model-based technique to post-process predictions and suppress high-frequency noise \cite{lippePDERefinerAchievingAccurate2023}. Transfer learning, potentially combined with selective fine-tuning on longer trajectories, has also demonstrated improved robustness \cite{xuTransferLearningEnhanced2023}. Other studies focus directly on the stability properties of the learned dynamics, exploring modifications to FNO architectures aimed at improving autoregressive stability \cite{mccabeStabilityAutoregressiveNeural2023}. Furthermore, techniques inspired by advanced sequence modeling have been investigated. COAST utilizes a causally-aware transformer architecture with adaptive time-stepping capabilities \cite{wuCOASTIntelligentTimeAdaptive2025}, while recent work suggests that state-space models like Mamba might be adept at capturing long-range temporal dependencies \cite{huStatespaceModelsAre2025}. Although these approaches show promise, they often entail increased model complexity, additional post-processing steps, or computationally intensive fine-tuning stages, and many still rely fundamentally on an initial teacher-forced training phase.
\subsection{Our Perspective: Recurrent Training for Improved Extrapolation}

Our approach differs from methods that emphasize post-hoc corrections or rely on complex sequence architectures; instead, we directly address the fundamental train-test mismatch inherent in the standard teacher-forcing paradigm. We propose a recurrent training strategy that explicitly aligns the training dynamics with the autoregressive nature of inference. The central idea is to recursively apply the operator to its own outputs during training (within the interval $[0, T]$), thereby exposing the model to its own prediction errors and the resulting distribution shift induced by multi-step rollouts. This exposure encourages the model to learn dynamics that are more robust to perturbations encountered in long-term forecasting, promoting stability and accuracy throughout the autoregressive process. Moreover, this training paradigm offers a more natural foundation for theoretical analysis of stability and error propagation within the neural operator framework. 
\begin{figure}[htbp]
  \centering
  \subfigure{
      \begin{adjustbox}{valign=m}
          \includegraphics[height=0.2\textheight, trim=44 63 43 62, clip]{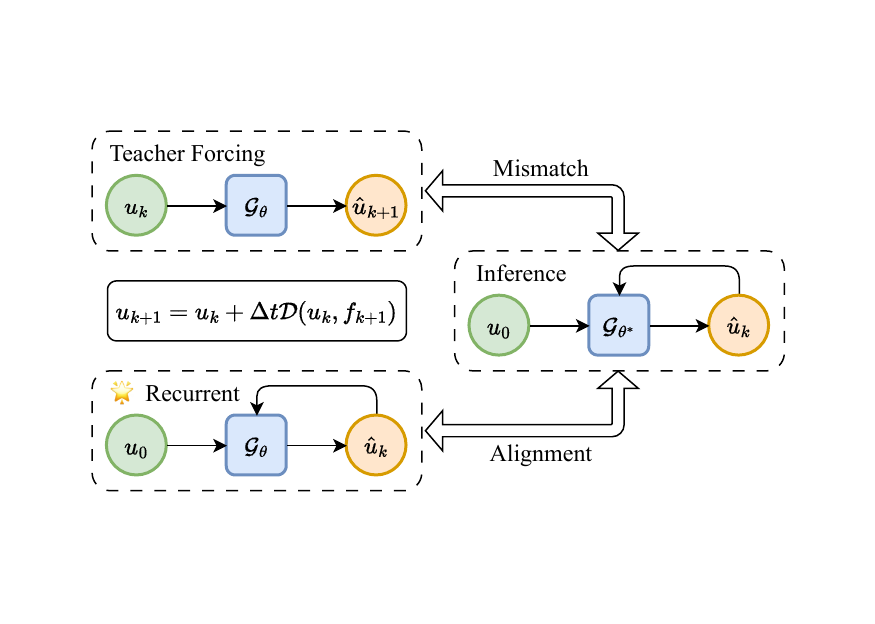}
      \end{adjustbox}
  }
  \hspace{4pt}
  \subfigure{
      \begin{adjustbox}{valign=m}
          \includegraphics[height=0.23\textheight]{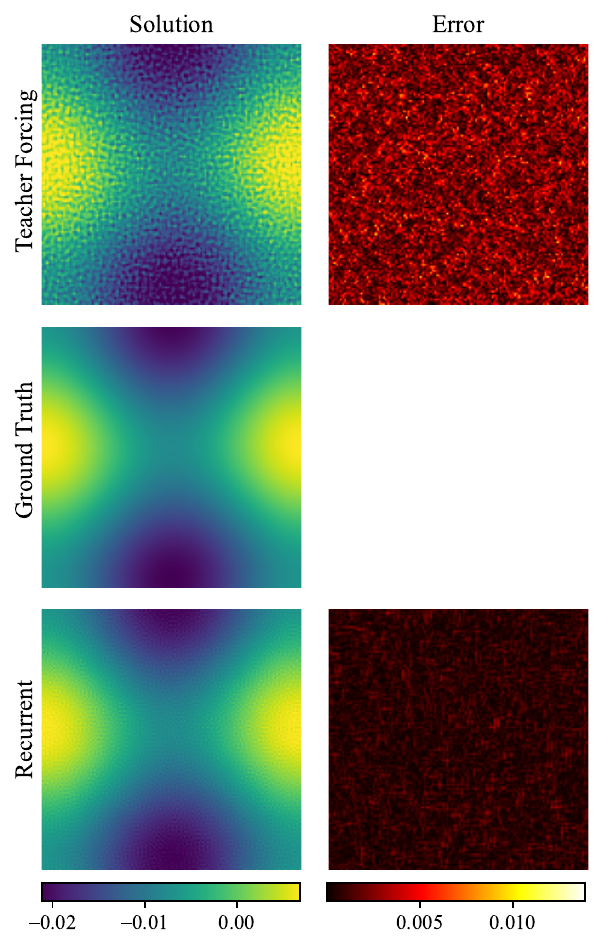}
      \end{adjustbox}
  }
  \caption{Recurrent Neural Operators. (Left) RNO aligns training with autoregressive inference, unlike teacher forcing which causes a train-test mismatch. (Right) Qualitative comparison on the transient heat conduction problem: teacher-forcing (top), ground truth (middle), and RNO (bottom) predictions with corresponding absolute errors. RNO shows improved long-term stability.}
\end{figure}

\section{Methodology}
\label{sec:methodology}
This section details the architecture of the neural operators employed in our study, introduces the proposed recurrent training framework, and presents a theoretical analysis comparing error propagation under recurrent versus conventional training strategies.

\subsection{Deep Neural Operator Architecture}

Following established frameworks \cite{heMgNOEfficientParameterization2024, kovachkiNeuralOperatorLearning2023, liFourierNeuralOperator2021}, we define the structure of a deep neural operator with $L$ hidden layers. Let the input be $g^0 = f \in \mathbb{F}$. The propagation through the network is defined as:
\begin{equation}
    \left\{
         \begin{aligned}
            g^0 &= f \in \mathbb{F}, \\
            g^l &= \sigma \left( \mathcal{K}^l g^{l-1} + b^l \right), && l = 1, \dots, L, \\
            \mathcal{G}_\theta(f) &= \mathcal{Q} g^{L}, && 
         \end{aligned}   
    \right.
\end{equation}
where $g^l$ represents the hidden representation in the $l$-th layer, residing in a function space $\mathbb{U}^{n_l}$ (the Cartesian product of $n_l$ copies of a base Banach space $\mathbb{U}$). The operator $\mathcal{K}^l$ denotes the primary linear operator of the $l$-th layer, mapping from $\mathbb{U}^{n_{l-1}}$ to $\mathbb{U}^{n_l}$ (with $\mathbb{U}^{n_0} = \mathbb{F}$). For $l=1$, $\mathcal{K}^1$ maps from $\mathbb{F}$ to $\mathbb{U}^{n_1}$. The term $b^l \in \mathbb{U}^{n_l}$ represents a bias function, and $\sigma$ is an element-wise nonlinear activation function. Finally, $\mathcal{Q}$ is an output transformation mapping the last hidden representation $g^L \in \mathbb{U}^{n_L}$ to the desired output space $\mathbb{U}$. We denote the entire neural operator parameterized by $\theta$ as $\mathcal{G}_\theta$. For simplicity, we assume all hidden layers have the same width, $n_l = n$, for $l=1, \dots, L$.

The bias function $b^l = (b^l_1, \dots, b^l_n) \in \mathbb{U}^n$ is often parameterized as a learnable affine transformation of the previous layer's output $g^{l-1} = (g^{l-1}_1, \dots, g^{l-1}_n)$:
\begin{equation}\label{eq:NO-LinearCombination}
    \mathbb{U} \ni b^l_i = \sum_{j=1}^n \alpha_{ij}^l g^{l-1}_j + \beta_i^l, \quad i = 1, \dots, n, \quad l = 1, \dots, L,
\end{equation}
where $\alpha_{ij}^l$ and $\beta_i^l$ are learnable scalar parameters.

Different choices for parameterizing the linear operators $\mathcal{K}^l$ yield various neural operator architectures. In this work, we primarily consider two state-of-the-art approaches: the FNO \cite{liFourierNeuralOperator2021}, which parameterizes $\mathcal{K}^l$ via the Fourier domain, and the MgNO \cite{heMgNOEfficientParameterization2024}, which employs multigrid-inspired structures. Formal definitions of these operators are provided in Appendix~\ref{sec:formal representation of neural operators}.

\subsection{Recurrent Operator Learning Framework}

To study the solution trajectory of the evolution equation \eqref{eq:base_problem} over a time interval $[0, T]$, a common strategy is to discretize time using numerical schemes such as the Euler method. Given a time step $\Delta t = T/N$, the discretized form reads:
\begin{equation}\label{eq:EulerScheme}
    u_{n+1} = u_n + \Delta t \, \mathcal{D}(u_n, f_{n+1}), \quad n = 0, \dots, N-1,
\end{equation}
where $u_n$ approximates the solution $u(n\Delta t, \cdot)$ and $f_{n+1}$ denotes external inputs, parameters, or forcing terms relevant to the time interval $[n\Delta t, (n+1)\Delta t]$. The learning objective is to approximate the dynamics operator $\mathcal{D}$—or more practically, the one-step solution map—using a neural operator $\mathcal{G}_\theta$ trained to emulate this temporal evolution.

The standard training paradigm, often referred to as \textbf{teacher forcing} or single-step prediction, trains the operator $\mathcal{G}_\theta$ to predict the state $u_{n+1}$ given the ground-truth state $u_n$ at the previous step:
\begin{equation}\label{eq:teacher_forcing_train}
    \min_\theta \sum_{n=0}^{N-1} \left\| u_n + \Delta t \mathcal{G}_\theta(u_n, f_{n+1}) - u_{n+1} \right\|,
\end{equation}
where $\|\cdot\|$ is a suitable norm. While computationally efficient, this approach creates a mismatch: during training, $\mathcal{G}_\theta$ always receives ground-truth inputs ($u_k$), whereas during inference (autoregressive rollout for $t>T$ or when ground truth is unavailable), it receives its own previous predictions ($\hat{u}_k$). This train-test mismatch leads to compounding errors, particularly for long-term predictions.

To address this fundamental limitation, we propose the \textbf{Recurrent Neural Operator} training framework. Instead of using ground-truth states at each step, the RNO approach performs an autoregressive rollout during training within the horizon $[0, T]$. Let $\mathcal{G}_\theta$ be the neural operator intended to approximate the dynamics update (i.e., approximating $\mathcal{D}$ in Eq.~\eqref{eq:EulerScheme}). The RNO iteratively computes predicted states:
\begin{equation}\label{eq:rno_iteration}
    \left\{
        \begin{aligned}
            \hat{u}_0 &= u_0, \\ 
            \hat{u}_{n+1} &= \hat{u}_n + \Delta t \mathcal{G}_\theta (\hat{u}_n, f_{n+1}), \quad n = 0, 1, \dots, N-1.
        \end{aligned}
    \right.
\end{equation}
The training objective then minimizes the accumulated error over the entire trajectory within the training horizon:
\begin{equation}\label{eq:rno_train}
   \min_\theta \sum_{n=1}^{N} \left\|\hat{u}_n - u_n\right\|. 
\end{equation}
Crucially, the same operator $\mathcal{G}_\theta$ is applied recurrently at each step $n$. By optimizing the parameters $\theta$ based on the quality of the entire rollout $\{\hat{u}_n\}_{n=1}^N$, the RNO framework forces the operator $\mathcal{G}_\theta$ to learn dynamics that are robust to its own predictive errors. This inherent alignment between the training procedure and the autoregressive nature of inference is hypothesized to mitigate the distribution shift problem and significantly improve the stability and accuracy of long-term extrapolations, as supported by our theoretical analysis in the following subsection.

\subsection{Theoretical Comparison of Error Propagation}
\label{sec:theoretical_result}
To provide theoretical insight into the benefits of recurrent training, we analyze the propagation of errors over a finite time horizon $[0, T]$ for both teacher forcing and recurrent training strategies, using a neural operator $\mathcal{G}_\theta$ to approximate the dynamics update in the Euler scheme \eqref{eq:EulerScheme}. The following theorem contrasts the error accumulation under the two training paradigms. The proof is deferred to Appendix~\ref{sec:proof}.

\begin{theorem}[Error Bounds for Teacher Forcing vs. Recurrent Training]\label{theorem:error_estimate_comparison}
Let $u_n$ be the exact solution sequence following of the problem \eqref{eq:base_problem}, and let $\hat{u}_n^{TF}$ and $\hat{u}_n^{RNO}$ be the approximate solutions generated using an operator $\mathcal{G}_\theta$ trained via teacher forcing and recurrent training, respectively. Assume the learned operator $\mathcal{G}_\theta$ satisfies the universal approximation theorem with the error bounded $\epsilon$.

(i) \textbf{Teacher Forcing (TF):} If $\mathcal{G}_\theta$ is trained via \eqref{eq:teacher_forcing_train}, the error during autoregressive rollout satisfies:
\begin{equation}
    \max_{0 \leq n \leq N} \left\| \hat{u}_n^{TF} - u_n \right\| \leq e^{C T} \left\| \hat{u}_0 - u_0 \right\| + \frac{\left(e^{C T} - 1\right)}{C} \left(\epsilon + O(\Delta t)\right),
\end{equation}
where $C > 0$ is the Lipschitz constant of the neural operator $\mathcal{G}_\theta$ with respect to its input.

(ii) \textbf{Recurrent Training (RNO):} If $\mathcal{G}_\theta$ is trained via \eqref{eq:rno_train}, the error during autoregressive rollout satisfies:
\begin{equation}
    \max_{0 \leq n \leq N} \left\| \hat{u}_n^{RNO} - u_n \right\| \leq \left\| \hat{u}_0 - u_0 \right\| + T \left(\epsilon + O(\Delta t)\right).
\end{equation}
\end{theorem}

Theorem \ref{theorem:error_estimate_comparison} highlights a crucial theoretical distinction between the two training paradigms regarding error accumulation.
The bound associated with teacher forcing (Part i) typically exhibits exponential dependence on the time horizon $T$. This reflects the potential for rapid error amplification during autoregressive rollout, stemming from the mismatch between training conditions and the sequential prediction task. This theoretical amplification aligns with the instabilities often observed empirically when using teacher-forced models for long-term extrapolation.

In contrast, the bound associated with recurrent training (Part ii), under the assumption that the training successfully enforces stability, shows error growth that is potentially much slower, ideally approaching linear dependence on $T$. By optimizing the operator based on its performance in autoregressive sequences during training, the RNO framework directly encourages the learning of dynamics that control error propagation. The resulting bound suggests that errors accumulate additively rather than multiplicatively, providing theoretical support for the hypothesis that recurrent training leads to significantly improved long-term stability and extrapolation accuracy.

Both bounds depend on the initial error $\| \hat{u}_0 - u_0 \|$, the operator's intrinsic approximation accuracy $\epsilon$, and the time discretization error $O(\Delta t)$. However, the qualitative difference in their dependence on the time horizon $T$ ($e^{CT}$ vs. $T$) is the key takeaway, strongly favoring the recurrent training approach for tasks requiring reliable long-term prediction. While practical estimation of constants like $C$ and $\epsilon$ remains challenging, the theorem provides a valuable conceptual framework for understanding why recurrent training is advantageous for learning stable dynamics from time-series data.

\section{Experiments and Results}
\label{sec:experiments}

This section evaluates the performance of the proposed RNO training framework against baseline methods across several benchmark problems derived from time-dependent PDEs.

\subsection{Overview of Benchmarks and Baselines}

To assess the effectiveness and generalization capabilities of the RNO approach, we conducted comprehensive experiments on a diverse set of benchmark problems. These benchmarks are based on various PDEs commonly encountered in scientific modeling: the transient heat conduction equation, the Allen–Cahn equation, the Cahn–Hilliard equation, and the incompressible Navier–Stokes equations. 
Detailed specifications of these benchmark PDEs are provided in Appendix~\ref{section:benchmarkPDEs}.

We compare two primary training strategies for neural operators: the teacher forcing approach \eqref{eq:teacher_forcing_train} and the recurrent training approach \eqref{eq:rno_train}. For both strategies, we evaluate two underlying neural operator architectures: FNO \cite{liFourierNeuralOperator2021} and MgNO \cite{heMgNOEfficientParameterization2024}. Additionally, we include PDE-Refiner \cite{lippePDERefinerAchievingAccurate2023} as a strong baseline representing methods that employ post-hoc correction steps to improve long-term stability.

\subsection{Basic Rollout Performance}

In this primary experiment, we evaluate the ability of different models to predict the system's evolution beyond the training horizon. Each model, trained on data up to $n=10$ steps, is used to generate autoregressive predictions up to $n=50$ steps. We measure performance using the mean relative $L^2$ error between the prediction and the ground truth, averaged over a test set of initial conditions. Table~\ref{tab:basic_rollout} presents the mean relative $L^2$ errors, with the mean result among three independent training runs reported for each setting. Values in parentheses represent one standard deviation. We report errors at $n=5$ (short-term accuracy within the training horizon distribution) and $n=50$ (long-term extrapolation accuracy and stability). Lower values indicate better performance.

\begin{table}[htbp]
    \centering
    \setlength{\tabcolsep}{5pt}
    \caption{Comparison of mean relative $L^2$ errors for basic rollout prediction at $n=5$ and $n=50$ steps. Models were trained up to $n=10$. 'tf-' denotes teacher forcing, 'r-' denotes recurrent training. Lower is better. Values in parentheses denote one standard deviation over three independent runs. Parameter counts (Param) are in millions.}
    \label{tab:basic_rollout}
    \begin{tabular}{c c c c c c c c c}
    \toprule
    \multirow[b]{2}{*}{\makecell[c]{Model\\ (Param)}} & \multicolumn{2}{c}{\textbf{Heat}} & \multicolumn{2}{c}{\textbf{Allen–Cahn}} & \multicolumn{2}{c}{\textbf{Cahn–Hilliard}} & \multicolumn{2}{c}{\textbf{Navier–Stokes}} \\
    \cmidrule(r){2-3}
    \cmidrule(r){4-5}
    \cmidrule(r){6-7}
    \cmidrule(r){8-9}
     &   $n=5$ & $n=50$ & $n=5$ & $n=50$ & $n=5$ & $n=50$ & $n=5$ & $n=50$ \\
    \midrule
    \makecell{Refiner \vspace{-2pt}\\ {\scriptsize ($9.1$M)}} &   \makecell{3.7e-03 \vspace{-2pt}\\ {\scriptsize($\pm$3.2e-04)}} & \makecell{5.1e-02 \vspace{-2pt}\\ {\scriptsize($\pm$3.8e-02)}} & \makecell{1.7e-03 \vspace{-2pt}\\ {\scriptsize($\pm$2.2e-04)}} & \makecell{2.9e-02 \vspace{-2pt}\\ {\scriptsize($\pm$6.7e-03)}} & \makecell{1.5e-03 \vspace{-2pt}\\ {\scriptsize($\pm$2.4e-04)}} & \makecell{6.1e-03 \vspace{-2pt}\\ {\scriptsize($\pm$4.5e-04)}} & \makecell{3.8e-02 \vspace{-2pt}\\ {\scriptsize($\pm$4.8e-02)}} & \makecell{1.4e-01 \vspace{-2pt}\\ {\scriptsize($\pm$1.8e-01)}} \\
    \midrule
    \makecell{tf-FNO\vspace{-2pt}\\ {\scriptsize ($1.1$M)}} & \makecell{3.9e-02 \vspace{-2pt}\\ {\scriptsize ($\pm$8.9e-04)}} & \makecell{1.8e-01 \vspace{-2pt}\\ {\scriptsize ($\pm$1.3e-01)}} & \makecell{1.0e-02 \vspace{-2pt}\\ {\scriptsize ($\pm$8.6e-04)}} & \makecell{3.3e-02 \vspace{-2pt}\\ {\scriptsize ($\pm$5.4e-03)}} & \makecell{3.7e-02 \vspace{-2pt}\\ {\scriptsize ($\pm$5.3e-03)}} & \makecell{1.6e-01 \vspace{-2pt}\\ {\scriptsize ($\pm$7.0e-03)}} & \makecell{5.7e-02 \vspace{-2pt}\\ {\scriptsize ($\pm$2.6e-03)}} & \makecell{2.4e\raisebox{0.1ex}{\scalebox{0.7}{+}}02 \vspace{-2pt}\\ {\scriptsize ($\pm$3.0e+02)}} \vspace{2pt}\\
    
    \makecell{tf-MgNO\vspace{-2pt}\\ {\scriptsize ($0.8$M)}} & \makecell{7.4e-03 \vspace{-2pt}\\ {\scriptsize ($\pm$1.9e-04)}} & \makecell{2.9e-02 \vspace{-2pt}\\ {\scriptsize ($\pm$2.2e-03)}} & \makecell{6.0e-03 \vspace{-2pt}\\ {\scriptsize ($\pm$8.5e-04)}} & \makecell{5.8e-03 \vspace{-2pt}\\ {\scriptsize ($\pm$1.1e-03)}} & \makecell{1.3e-03 \vspace{-2pt}\\ {\scriptsize ($\pm$2.4e-04)}} & \makecell{3.4e-03 \vspace{-2pt}\\ {\scriptsize ($\pm$8.4e-04)}} & \makecell{2.2e-02 \vspace{-2pt}\\ {\scriptsize ($\pm$1.7e-02)}} & \makecell{1.4e-01 \vspace{-2pt}\\ {\scriptsize ($\pm$9.5e-02)}} \\

    \midrule
    \makecell{r-FNO\vspace{-2pt}\\ {\scriptsize ($1.1$M)}} & \makecell{3.2e-02 \vspace{-2pt}\\ {\scriptsize ($\pm$3.1e-03)}} & \makecell{1.2e-01 \vspace{-2pt}\\ {\scriptsize ($\pm$6.1e-02)}} & \makecell{9.7e-03 \vspace{-2pt}\\ {\scriptsize ($\pm$1.3e-03)}} & \makecell{3.0e-02 \vspace{-2pt}\\ {\scriptsize ($\pm$7.4e-03)}} & \makecell{3.2e-02 \vspace{-2pt}\\ {\scriptsize ($\pm$5.6e-03)}} & \makecell{2.1e-01 \vspace{-2pt}\\ {\scriptsize ($\pm$3.2e-02)}} & \makecell{5.5e-02 \vspace{-2pt}\\ {\scriptsize ($\pm$2.9e-03)}} & \makecell{3.7e-01 \vspace{-2pt}\\ {\scriptsize ($\pm$3.5e-01)}} \vspace{2pt}\\

    \makecell{\textbf{r-MgNO}\vspace{-2pt}\\ {\scriptsize ($0.8$M)}}  & \makecell{\textbf{3.6e-03} \vspace{-2pt}\\ {\scriptsize ($\pm$3.3e-04)}} & \makecell{\textbf{1.1e-02} \vspace{-2pt}\\ {\scriptsize ($\pm$1.4e-03)}} & \makecell{\textbf{3.1e-04} \vspace{-2pt}\\ {\scriptsize ($\pm$1.2e-04)}} & \makecell{\textbf{6.0e-04} \vspace{-2pt}\\ {\scriptsize ($\pm$3.8e-05)}} & \makecell{\textbf{1.2e-03} \vspace{-2pt}\\ {\scriptsize ($\pm$2.4e-04)}} & \makecell{\textbf{2.9e-03} \vspace{-2pt}\\ {\scriptsize ($\pm$6.2e-04)}} & \makecell{\textbf{3.4e-03} \vspace{-2pt}\\ {\scriptsize ($\pm$6.6e-04)}} & \makecell{\textbf{8.9e-02} \vspace{-2pt}\\ {\scriptsize ($\pm$5.4e-02)}} \\

    \bottomrule
    \end{tabular}
\end{table}

The basic rollout experiments, summarized in Table~\ref{tab:basic_rollout}, underscore the substantial benefits of adopting a recurrent training strategy, especially when applied to the MgNO architecture. 
First, the recurrent formulation significantly improves long-term prediction stability in conjunction with MgNO (r-MgNO), reducing the mean relative $L^2$ error at $n=50$ by approximately $9.1\%$ to $87.2\%$ across all benchmark PDEs compared to the standard teacher-forcing method (tf-MgNO).
Second, the results emphasize the importance of architectural compatibility: while MgNO benefits substantially from recurrent training, the FNO architecture shows inconsistent outcomes (r-FNO vs. tf-FNO), suggesting that MgNO’s structure may be inherently better suited for learning stable dynamics through recurrent state propagation. 
Furthermore, r-MgNO achieves competitive performance with fewer parameters than the PDE-Refiner, demonstrating the efficiency of the proposed training approach. Overall, these findings highlight how the RNO strategy unlocks the full potential of MgNO for robust long-horizon forecasting by aligning training with autoregressive inference, while preserving parameter efficiency.

\subsection{Long-Horizon Rollout Performance}
\begin{figure}[htbp]
    \centering
    \subfigure[Rollout up to $n=50$]{
        \includegraphics[width=0.45\textwidth]{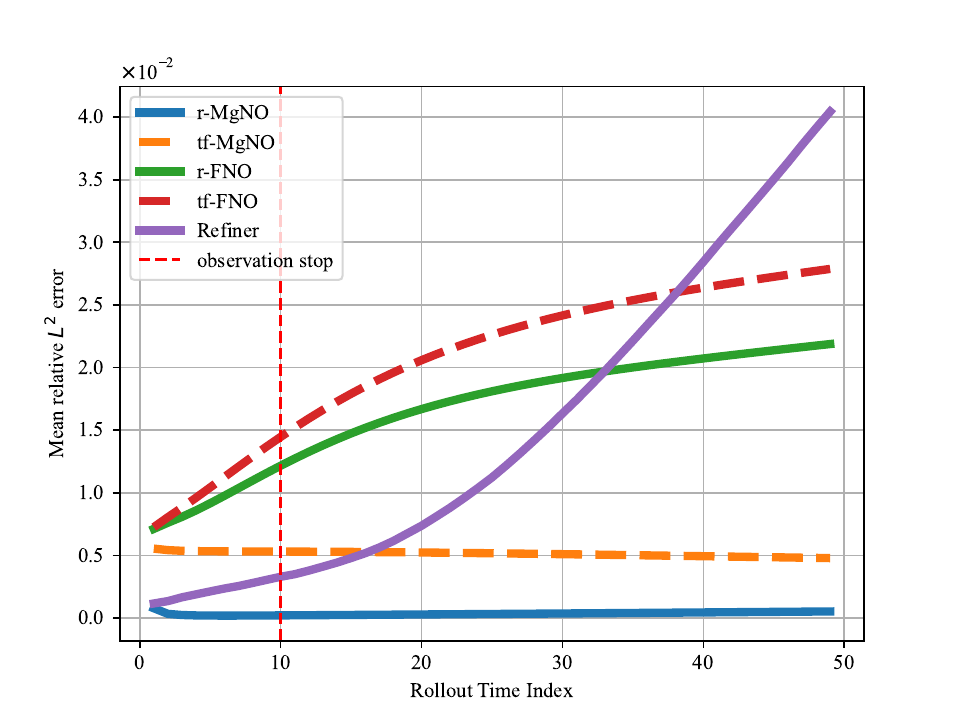}
        \label{fig:horizon50}
    }
    \hfill
    \subfigure[Rollout up to $n=500$]{
        \includegraphics[width=0.45\textwidth]{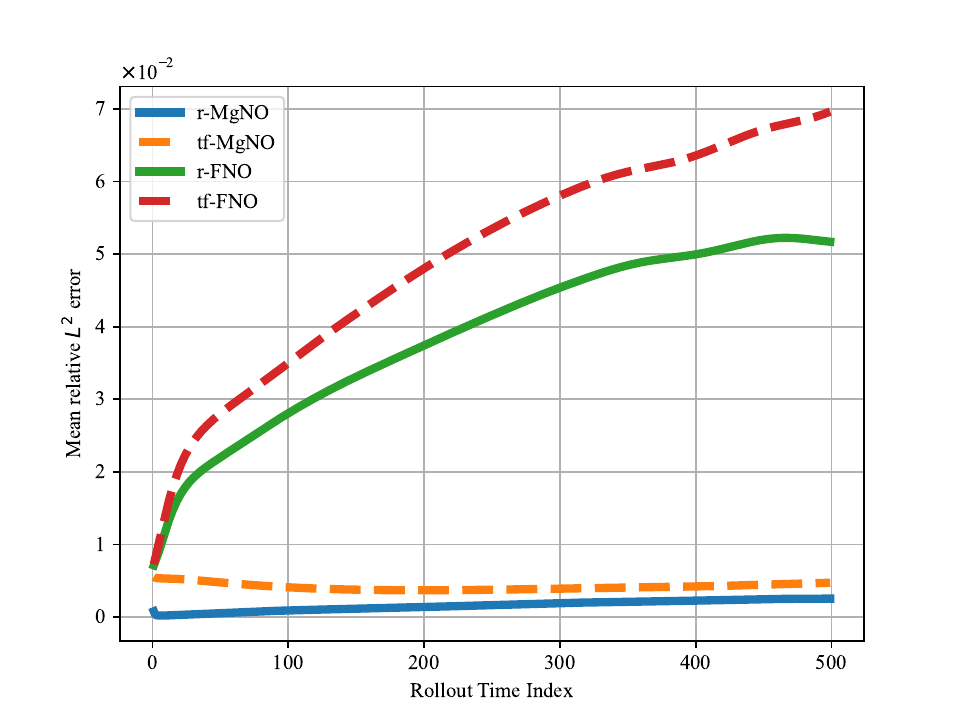}
        \label{fig:horizon500}
    }
    \caption{Comparison of long-term rollout stability on the Allen–Cahn equation. Both panels show the mean relative $L^2$ error versus rollout time steps. Models were trained for $n=10$ time steps. (a) Performance up to $n=50$ time steps. (b) Performance over an extremely long horizon up to $n=500$ time steps.}
    \label{fig:long_term_rollout}
\end{figure}

To comprehensively evaluate the long-term stability and predictive accuracy of different training strategies and operator architectures, we conducted rollout experiments on the Allen–Cahn benchmark. Figure~\ref{fig:long_term_rollout} illustrates the results, revealing that the synergistic application of MgNO and cyclical training yielded minimal error and the most stable progression of error accumulation. This indicates that RNO represents a potent methodology for long-horizon forecasting in this context.

\subsection{Ablation Study}
This ablation study is designed to systematically evaluate the robustness and generalization ability of the proposed recurrent training method. Using the Allen-Cahn problem as a representative example, we analyze the effect of the observation time length used during training, the time step size ($\Delta t$) employed for discretization and the data size $N^{data}$. The goal is to understand how these factors influence the model's performance and stability, particularly its long-term extrapolation capabilities achieved through recurrent training. We use the r-MgNO model configuration based on its superior performance shown earlier.

\begin{figure}[htbp]
    \centering
    \subfigure[Length of observation time]{
        \includegraphics[width=0.45\textwidth]{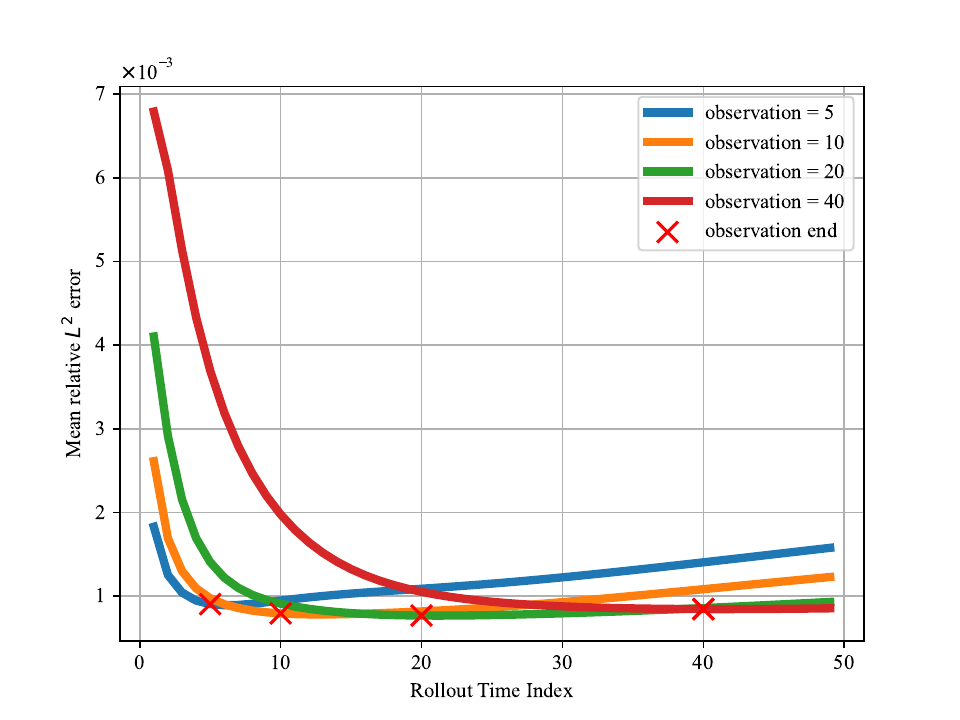}
        \label{fig:ablation_obs_len}
    }
    \hfill \subfigure[Time step ($\Delta t$)]{
        \includegraphics[width=0.45\textwidth]{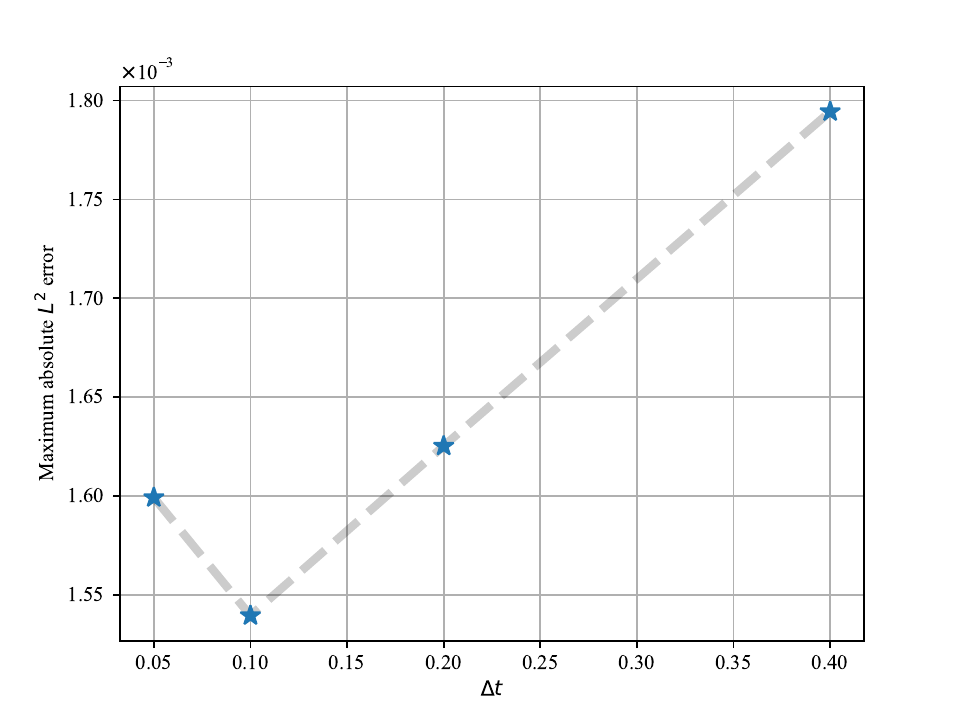}
        \label{fig:ablation_time_step}
    }
    \caption{Ablation study on the Allen-Cahn equation using the r-MgNO model. (a) Effect of varying the number of observation steps during training. (b) Effect of varying the time step size ($\Delta t$).}
    \label{fig:ablation}
\end{figure}

\textbf{Effect of Observation Time Length:} Figure \ref{fig:ablation_obs_len} illustrates the impact of varying the number of steps observed during training on the model's predictive accuracy. For long-term extrapolation ($n=50$), the error significantly decreases as the observation length increases from $5$ to $40$ steps, indicating that observing longer trajectories during recurrent training helps the model learn the dynamics more effectively for stable prediction. However, another phenomenon is that if the observation time is too long, the error at the initial moment will be larger. This reflects a drawback of recurrent training, namely that due to the recurrent structure leading to vanishing gradients, the model's ability to learn the initial state is weaker.

\textbf{Effect of Time Step Size:} We train the r-MgNO model within the fixed time horizon $[0, 2]$ and vary the time step size $\Delta t$. Figure \ref{fig:ablation_time_step} shows the maximum relative error over the time horizon $[0, 10]$. When $\Delta t$ decreases from $0.4$ to $0.1$, the error shows a linearly decreasing trend, which is consistent with the content of Theorem \ref{theorem:error_estimate_comparison}. However, when the error further decreases to $0.05$, the error increases. This might be because a smaller $\Delta t$ leads to training difficulties, resulting in a larger approximation error for the neural operator.

\textbf{Effect of Data Size:} Based on the data presented in Table~\ref{tab:ablation_data_size}, the r-MgNO method demonstrates superior performance, consistently achieving lower mean relative $L^2$ errors across varying data sizes and exhibiting a significantly higher order of convergence, indicating faster error reduction with increased training data. 
The order is computed as $\log(e_{i+1} / e_i) / \log(N_{i+1}^{data} / N_i^{data})$, where $e_i$ and $e_{i+1}$ are the mean relative $L^2$ errors corresponding to training data sizes $N_i^{data}$ and $N_{i+1}^{data}$, respectively. A higher order indicates faster error decay with increasing data.

\begin{table}[htbp]
    \centering
    \caption{Effect of training data size on mean relative L2 error for extrapolation ($n=50$). This order indicates how quickly the error decreases as the amount of data increases, reflecting the efficiency of accuracy improvement in the algorithm.}
    \label{tab:ablation_data_size}
    \begin{tabular}{ccccc}
        \toprule
        Data size & tf-MgNO & Order & r-MgNO & Order \\
        \midrule
        $250$ & 1.0e-02 & - & 8.2e-03 & - \\
        $500$ & 8.8e-03 & 0.217 & 3.5e-03 & 1.228 \\
        $1000$ & 4.7e-03 & 0.888 & 6.1e-04 & 2.525 \\
        $2000$ & 4.7e-03 & 0.015 & 5.7e-04 & 0.099 \\
        \bottomrule
        \end{tabular}
\end{table}

\subsection{Training Cost}

Table~\ref{tab:training-cost} summarizes the training costs of various models on the Allen-Cahn benchmark. We observe that the recurrent model r-MgNO incurs higher computational overhead per epoch and consumes more GPU memory compared to tf-MgNO and Refiner. This is expected due to the recurrent nature of r-MgNO, which processes multiple time steps during training. Nevertheless, the training time and memory usage remain within practical limits, especially when using distributed training with DDP.

\begin{table}[htbp]
    \centering
    \caption{Comparison of computational costs for different models and training setups. DDP indicates training on four NVIDIA A100 GPUs using Distributed Data Parallel (DDP). Time is wall-clock time per epoch. Memory is peak GPU memory usage per GPU.}
    \label{tab:training-cost}
    \begin{tabular}{lccc}
    \toprule
    \textbf{Model} & \textbf{Param (M)} & \textbf{Time per Epoch (s)} & \textbf{Memory per GPU (GB)} \\
    \midrule
    Refiner          & $9.1$  & $44.0$  & $4.3$  \\
    Refiner (DDP)         & $9.1$  & $13.0$  & $4.5$  \\
    tf-MgNO          & $0.8$  & $4.0$  & $5.7$  \\
    tf-MgNO (DDP)    & $0.8$  & $1.1$  & $5.8$  \\
    r-MgNO           & $0.8$  & $23.6$  & $34.5$ \\
    r-MgNO (DDP)     & $0.8$  & $5.8$  & $34.6$ \\
    \bottomrule
    \end{tabular}
\end{table} 
\section{Conclusions}
\label{sec:conclusions}
In this work, we addressed error accumulation in long-term autoregressive prediction for neural operators on time-dependent PDEs by proposing the Recurrent Neural Operator (RNO), which aligns training with inference through recursive application.

Our main contributions are twofold. First, we theoretically proved that recurrent training reduces worst-case error growth from exponential to linear over the prediction horizon. Second, empirical results demonstrate that RNOs (especially r-MgNO) surpass teacher-forced methods in long-term accuracy and stability, even with fewer parameters than post-hoc correction approaches. These advances address a core limitation in neural operator forecasting, enhancing the reliability of data-driven dynamical models.

However, recurrent training incurs higher per-epoch costs and may struggle with extremely long sequences due to vanishing gradients. Future directions include developing efficient training algorithms (e.g., gradient clipping or sparse recurrence), designing architectures for stability, and applying RNOs to real-world systems such as climate modeling or turbulent flow prediction.

\newpage
\bibliographystyle{plain}

\bibliography{RNO}

\newpage
\appendix
\label{sec:appendices}
\section{Formal Representation of Neural Operators}
\label{sec:formal representation of neural operators}
\subsection{Fourier Neural Operator}
In FNO, the basic linear operator is defined through a process involving the Fourier transform, a linear combination of Fourier modes, and the inverse Fourier transform. 
\begin{equation*}
    \mathbb{U} \ni \left[ \mathcal{K}^l g^{l-1} \right]_i = \mathcal{F}^{-1} \left( \sum_{j=1}^n \gamma_{ij}^l \mathcal{F} \left(g^{l-1}_j\right)\right) ,\quad i = 1:n, \quad l = 1:L.
\end{equation*}
Here, $\gamma_{ij}^l$ are learnable parameters, $\mathcal{F}$ and $\mathcal{F}^{-1}$ represent the Fourier transform and its inverse, respectively. When the input is discrete, the Fourier transform can be approximated using the discrete Fourier transform. In the actual implementation, high-frequency modes are deliberately truncated to exclude noise. 

\subsection{Multigrid Neural Operator}
In MgNO, the basic linear operator is defined through a process involving a series of multigrid operations, which comprises three core components: residual correction iterations, restriction operations, and prolongation operations. These components are formally defined as follows:

The residual correction procedure performs feature extraction and smoothing on fixed-resolution grids. For a grid with resolution $a$, the $\kappa$-step iteration process is defined as:
\begin{equation*}
    \left(f^{(a)}, u_{\kappa}^{(a)}\right) = \mathcal{I}_\kappa^{(a)}\left(f^{(a)}, u^{(a)}\right) 
    \Longleftrightarrow
    \left\{
    \begin{aligned}
        u_{0}^{(a)} &= u^{(a)} \\
        u_{k}^{(a)} &= u_{k-1}^{(a)} + \mathcal{B}_k \ast \left( f^{(a)} - \mathcal{A}_k \ast u_{k-1}^{(a)} \right),
    \end{aligned}
    \right.
\end{equation*}
where, superscript $(a)$ denotes grid resolution level, $\kappa \in \mathbb{N}^+$ specifies the number of smoothing iterations. $\mathcal{A}_k, \mathcal{B}_k \in \mathbb{R}^{d \times d}$ are learnable convolution kernels, and $\ast$ represents standard convolution operation.

The restriction operator transfers residual information from fine grid $a$ to coarse grid $b$:
\begin{equation*}
    \left(f^{(b)}, u^{(b)}\right) = \mathcal{R}_a^b\left(f^{(a)},u^{(a)}\right)
    \Longleftrightarrow
    u^{(b)} = 0,\;
    f^{(b)} = \mathcal{R} \ast_2 \left( f^{(a)} - \mathcal{A} \ast u^{(a)} \right),
\end{equation*}
where, $\ast_2$ denotes strided convolution with stride $2$, $\mathcal{R} \in \mathbb{R}^{d \times d}$ is a learnable restriction kernel.

The prolongation operator transfers corrections from coarse grid $b$ back to fine grid $a$:
\begin{equation*}
    \left(f^{(a)}, \hat{u}^{(a)}\right) = \mathcal{P}_b^a\left(f^{(b)},u^{(b)}\right)
    \Longleftrightarrow 
    \hat{u}^{(a)} = u^{(a)} + \mathcal{P} \ast^2 u^{(b)},
\end{equation*}
where, $\ast^2$ denotes transposed convolution with stride $2$, $\mathcal{P} \in \mathbb{R}^{d \times d}$ is a learnable prolongation kernel. $f^{(a)}$ and $u^{(a)}$ are inherited from the fine-grid. After the prolongation operation, Post-prolongation residual correction (post-smoothing) may be applied.

The complete multigrid operator $\mathcal{K}^l$ with $J$ restriction levels is defined through the V-cycle composition:
\begin{equation*}
    \left\{
        \begin{aligned}
            (g^{l-1}, \hat{u}^{(r_1)}_{\kappa_{2J-1}}) &= \mathcal{V}\text{-Cycle}\left(g^{l-1}, 0\right) ,\\
            \mathcal{K}^l g^{l-1} &= \hat{u}^{(r_1)} \in \mathbb{U}^n,
        \end{aligned}
    \right.
\end{equation*}
where the V-cycle process is structured as:
\begin{enumerate}
    \item Downward Path: $\mathcal{I}^{(r_1)}_{\kappa_1} \rightarrow \mathcal{R}_{r_1}^{r_2} \rightarrow \cdots \rightarrow \mathcal{R}_{r_{J-1}}^{r_J}$,
    \item Coarse Solve: $\mathcal{I}^{(r_J)}_{\kappa_J}$,
    \item Upward Path: $\mathcal{P}_{r_J}^{r_{J-1}} \rightarrow \cdots \rightarrow \mathcal{P}_{r_2}^{r_1} \rightarrow \mathcal{I}^{(r_1)}_{\kappa_{2J-1}}$,
\end{enumerate}
with resolution hierarchy $r_j$ where $r_1 > r_2 > \cdots > r_J$ denotes progressively coarser grids.

\section{Proof of Theorem~\ref{theorem:error_estimate_comparison}}
\label{sec:proof}
We begin by noting that the residual correction iterations, restriction operations, prolongation operations, and nonlinear activations involved in the MgNO are all bounded operators. Therefore, their composition remains bounded.

\begin{proposition}
    The multigrid neural operator is a bounded operator.
\end{proposition}

Furthermore, based on Theorem 3.1 in \cite{heMgNOEfficientParameterization2024}, the MgNO satisfies a universal approximation property.

\begin{lemma}
    For any continuous operator $\mathcal{O}^* : H^s(\Omega) \to H^r(\Omega)$ and any $\epsilon > 0$, there exists a neural operator $\mathcal{O}$ such that 
    \begin{equation*}
        \left\| \mathcal{O} u - \mathcal{O}^* u \right\| \leq \epsilon, \quad \forall u \in \mathcal{C},
    \end{equation*}
    where $\mathcal{C} \subset H^s(\Omega)$ is a compact set.
\end{lemma}

We now consider the application of MgNO to the time evolution of a dynamical system governed by a differential operator $\mathcal{D}$, with auxiliary conditions independent of $t$. Let the MgNO be used as the base model to approximate the dynamics.

Let $u_n = u(x, t_n)$ denote the exact solution at time step $t_n = n \Delta t$, and let $\hat{u}_n$ be the predicted solution by the neural operator $\mathcal{G}_\theta$. The true solution evolves according to a time-marching scheme (e.g., forward Euler for simplicity, matching the proof steps):
\begin{equation*}
    u_{n+1} = u_n + \Delta t \mathcal{D}(u_n, f) + O(\Delta t^2),
\end{equation*}
where $\mathcal{D}$ represents the exact differential operator governing the dynamics. The neural operator approximates this evolution:
\begin{equation*}
    \hat{u}_{n+1} = \hat{u}_n + \Delta t \mathcal{G}_\theta(\hat{u}_n, f).
\end{equation*}
The error at step $n+1$ is given by:
\begin{equation}
    \begin{aligned}
        e_{n+1} = \left\| \hat{u}_{n+1} - u_{n+1} \right\| &= \left\| \left(\hat{u}_n + \Delta t \mathcal{G}_\theta(\hat{u}_n, f)\right) - \left(u_n + \Delta t \mathcal{D}(u_n, f) + O(\Delta t^2)\right) \right\| \\
        &\leq \left\| \hat{u}_n - u_n \right\| + \Delta t \left\| \mathcal{G}_\theta(\hat{u}_n, f) - \mathcal{D}(u_n, f) \right\| + O(\Delta t^2) \\
        &= e_n + \Delta t \left\| \mathcal{G}_\theta(\hat{u}_n, f) - \mathcal{D}(u_n, f) \right\| + O(\Delta t^2). \label{eq:error_recurrence_base}
    \end{aligned}
\end{equation}

\subsection{Proof for Teacher Forcing (TF)}
In the teacher forcing setting, the operator $\mathcal{G}_\theta$ is trained to approximate $\mathcal{D}$ using the true state $u_n$ as input at each step during training. The error analysis proceeds by splitting the term $\left\| \mathcal{G}_\theta(\hat{u}_n, f) - \mathcal{D}(u_n, f) \right\|$ using the triangle inequality and adding and subtracting $\mathcal{G}_\theta(u_n, f)$:
\begin{align*}
\left\| \mathcal{G}_\theta(\hat{u}_n^{TF}, f) - \mathcal{D}(u_n, f) \right\| &\leq \left\| \mathcal{G}_\theta(\hat{u}_n^{TF}, f) - \mathcal{G}_\theta(u_n, f) \right\| + \left\| \mathcal{G}_\theta(u_n, f) - \mathcal{D}(u_n, f) \right\| \\
&\leq C \left\| \hat{u}_n^{TF} - u_n \right\| + \epsilon,
\end{align*}
where we used the Lipschitz continuity of $\mathcal{G}_\theta$ and the universal approximation property applied to the true state $u_n$.
    
Substituting this back into the error recurrence \eqref{eq:error_recurrence_base}:
\begin{equation*}
    e_{n+1}^{TF} \leq e_n^{TF} + \Delta t \left( C e_n^{TF} + \epsilon \right) + O(\Delta t^2) = (1 + C \Delta t) e_n^{TF} + \epsilon \Delta t + O(\Delta t^2).
\end{equation*}
For simplicity, let $e_n =e_n^{TF}= \left\| \hat{u}_n^{TF} - u_n \right\|$. By recursively applying this inequality:
\begin{align*}
e_n &\leq (1 + C \Delta t)^n e_0 + \left(\epsilon \Delta t + O(\Delta t^2)\right) \sum_{k=0}^{n-1} (1 + C \Delta t)^k \\
&= (1 + C \Delta t)^n e_0 + \left(\epsilon \Delta t + O(\Delta t^2)\right) \frac{(1 + C \Delta t)^n - 1}{(1 + C \Delta t) - 1} \\
&= (1 + C \Delta t)^n e_0 + \frac{\epsilon + O(\Delta t)}{C} \left((1 + C \Delta t)^n - 1\right).
\end{align*}
Let $N = T / \Delta t$ be the total number of time steps. Since $n \leq N$, we use the inequality $(1 + x/N)^N \leq e^x$, which implies $(1 + C \Delta t)^n \leq (1 + C T/N)^N \leq e^{C T}$. Therefore:
\begin{equation*}
    e_n \leq e^{C T} e_0 + \frac{e^{C T} - 1}{C} (\epsilon + O(\Delta t)).
\end{equation*}
Taking the maximum over $0 \leq n \leq N$:
\begin{equation*}
    \max_{0 \leq n \leq N} \left\| \hat{u}_n^{TF} - u_n \right\| \leq e^{C T} \left\| \hat{u}_0 - u_0 \right\| + \frac{e^{C T} - 1}{C} \left(\epsilon + O(\Delta t)\right).
\end{equation*}
This concludes the proof for part (i).
    
\subsection{Proof for Recurrent Training (RNO)}
In the recurrent training setting (assuming the training objective effectively minimizes the discrepancy term $\left\| \mathcal{G}_\theta(\hat{u}_n, f) - \mathcal{D}(u_n, f) \right\|$ over the rollout), we bound this term directly using the approximation error $\epsilon$. Note that the RNO loss is typically defined over the rollout, aiming to ensure $\mathcal{G}_\theta$ accurately predicts the next state even when given the previously predicted state $\hat{u}_n$ as input. The proof provided assumes this leads to the bound:
\begin{equation*}
    \left\| \mathcal{G}_\theta(\hat{u}_n^{RNO}, f) - \mathcal{D}(u_n, f) \right\| \leq \epsilon.
\end{equation*}
Substituting this into the error recurrence \eqref{eq:error_recurrence_base}:
\begin{equation*}
    e_{n+1}^{RNO} \leq e_n^{RNO} + \epsilon \Delta t + O(\Delta t^2).
\end{equation*}
Now, set $e_n = e_n^{RNO}= \left\| \hat{u}_n^{RNO} - u_n \right\|$. By recursively applying this inequality (summing from $k=0$ to $n-1$):
\begin{align*}
e_n &\leq e_0 + \sum_{k=0}^{n-1} \left( \epsilon \Delta t + O(\Delta t^2) \right) \\
&= e_0 + n (\epsilon \Delta t + O(\Delta t^2)).
\end{align*}
Let $N = T / \Delta t$. Since $n \leq N$, we have $n \Delta t \leq N \Delta t = T$. Thus:
\begin{equation*}
    e_n \leq e_0 + n \Delta t (\epsilon + O(\Delta t)) \leq e_0 + T (\epsilon + O(\Delta t)).
\end{equation*}
Taking the maximum over $0 \leq n \leq N$:
\begin{equation*}
    \max_{0 \leq n \leq N} \left\| \hat{u}_n^{RNO} - u_n \right\| \leq \left\| \hat{u}_0 - u_0 \right\| + T \left(\epsilon + O(\Delta t)\right).
\end{equation*}
This concludes the proof for part (ii).

\section{Benchmark PDE Problems}\label{section:benchmarkPDEs}

\subsection{Transient Heat Conduction Problem}
Consider the transient heat conduction equation defined on a spatial domain $\Omega = [0,1] \times [0,1]$ with periodic boundary conditions. The temperature field $u(x, t)$ evolves according to:
\begin{equation}
    \left\{
        \begin{aligned}
            &\frac{\partial u}{\partial t} = \alpha \Delta u + f, \quad x \in \Omega, \ t \geq 0, \\
            &u(x, 0) = u_0(x), \quad x \in \Omega.
        \end{aligned}
    \right.
\end{equation}
Here, $\alpha > 0$ is the thermal diffusivity, and $f$ denotes a source term representing internal heat generation. The function $u_0(x)$ represents the initial state of the field.

\subsection{Allen-Cahn Equation}
Consider the Allen-Cahn equation defined on a spatial domain $\Omega = [0,1] \times [0,1]$ with periodic boundary conditions. The equation describes the evolution of an order parameter $u(x,t)$ and is given by:
\begin{equation}
    \left\{
        \begin{aligned}
            &\frac{\partial u}{\partial t} = d_1 \Delta u + d_2 u(1 - u^2), \quad x \in \Omega, \ t \geq 0, \\
            &u(x, 0) = u_0(x), \quad x \in \Omega.
        \end{aligned}
    \right.
\end{equation}
Here, $d_1 > 0$ is the diffusion coefficient, and $d_2$ controls the nonlinearity responsible for phase separation. The function $u_0(x)$ represents the initial state of the field.

\subsection{Cahn-Hilliard Equation}
The Cahn-Hilliard equation is a fourth-order nonlinear PDE used to model phase separation and coarsening phenomena in binary mixtures:
\begin{equation}
    \left\{
        \begin{aligned}
            &\frac{\partial u}{\partial t} = \Delta w, \quad x \in \Omega, \ t \geq 0, \\
            &w = -d_1 \Delta u + d_2(u^3 - u), \\
            &u(x, 0) = u_0(x), \quad x \in \Omega.
        \end{aligned}
    \right.
\end{equation}
Here, $\Omega = [0,1] \times [0,1]$, $d_1 > 0$ is the interfacial energy coefficient, $d_2 > 0$ determines the strength of the nonlinear term, and $u_0(x)$ is the initial field configuration. The chemical potential $w$ couples the higher-order diffusion with the nonlinear term.

\subsection{Navier-Stokes Equation (Vorticity Form)}
The 2D incompressible Navier-Stokes equations in vorticity formulation are given by:
\begin{equation}
    \left\{
        \begin{aligned}
            &\frac{\partial w}{\partial t} + u \cdot \nabla w = \nu \Delta w + f(x), \quad x \in \Omega, \ t \geq 0, \\
            & \nabla \cdot u = 0\\
            &w(x, 0) = w_0(x), \quad x \in \Omega.
        \end{aligned}
    \right.
\end{equation}
Here, $\Omega = [0,1] \times [0,1]$, $w$ is the vorticity field, $u$ is the velocity field, $\nu > 0$ is the kinematic viscosity, and $f(x)$ represents external forcing. The velocity field $u$ is related to the vorticity $w$ via the Biot–Savart law.

\begin{table}[htbp]
    \centering
    \caption{Comparison of PDE models and their properties.}
    \begin{tabular}{cccccc}
    \toprule
    \textbf{Equation} & \textbf{Boundary} & \textbf{Source} & \textbf{Parameters} & \textbf{Conservation} & \textbf{Diffusion Type}\\
    \midrule
    Heat  & Periodic & Yes & $\alpha$ & -- &Linear \\
    AC & Periodic & No & $d_1, d_2$ & -- &Nonlinear\\
    CH & Periodic & No & $d_1, d_2$ & Mass &Fourth-order\\
    NS & Periodic & Yes & $\nu$ & Momentum & convection\\
    \bottomrule
    \end{tabular}
\end{table}

\subsection{Initial Condition Generation}

The initial condition of all problems are generated by sampling from a periodic Gaussian process with the covariance kernel:
\begin{equation}
k(\mathbf{x}, \mathbf{x}') = \exp\left( -\frac{1}{2\ell^2} \sum_{d=1}^2 \sin^2\left( \pi (x_d - x'_d) \right) \right),
\end{equation}
where $\ell$ is a length-scale parameter controlling spatial correlation. A covariance matrix is constructed over the discrete grid, with a small diagonal perturbation added for numerical stability. A Cholesky decomposition is performed, followed by sampling with a standard normal vector to produce the initial field, which is then reshaped into a two-dimensional form for use as the model's initial state.

\section{Experimental Setup}
To ensure the reproducibility of the results, the code for this paper has been open-sourced. Please visit: \href{https://anonymous.4open.science/r/RecurrentNeuralOperator-8B02/README.md}{Anonymous Github Link}.

\subsection{Data generation}
For all problems, the data is generated by computing reference solutions using traditional numerical methods, followed by appropriate spatial and temporal downsampling. The spatial resolution is $128 \times 128$ for unit square domain $\Omega = [0,1] \times [0,1]$, and the temporal resolution is $0.01$. The training set consists of $1000$ samples, and the test set contains $200$ independently generated samples. All data generation code will be made publicly available as open source.

\subsection{Model Architectures}
We consider three distinct model architectures for learning PDE dynamics:

\textbf{MgNO} adopts the architecture proposed in \cite{heMgNOEfficientParameterization2024}. This model utilizes $4$ layers, each with $32$ hidden channels, and processes single-channel input and output fields. A multigrid iteration strategy, specified by the pattern $[[1,0], [1,0], [1,0], [2,0], [2,0]]$, is incorporated across its layers.

\textbf{FNO} utilizes the architecture from \cite{liFourierNeuralOperator2023}. It comprises $4$ layers, each with $32$ hidden channels, and operates in the Fourier domain using 2D spectral modes of size $(16,16)$. To enhance capacity, it includes a projection layer with a channel expansion ratio of $2$.

\textbf{PDE-Refiner} is based on the architecture from PDE-Refiner \cite{lippePDERefinerAchievingAccurate2023}, configured with $16$ channels.

\subsection{Error Definition}

The mean relative $L^2$ error is defined as follows:

\begin{equation}
\text{Mean Relative } L^2 \text{ Error} = \frac{1}{N} \sum_{i=1}^{N} \frac{\| \hat{y}^{(i)} - y^{(i)} \|_2}{\| y^{(i)} \|_2},
\end{equation}

where $N$ is the total number of samples, $\hat{y}^{(i)}$ denotes the predicted output for the $i$-th sample, and $y^{(i)}$ is the corresponding ground truth. The relative error is computed per sample and then averaged over all samples.

\subsection{Hardware and Software Environment}
All experiments were conducted on a server running CentOS Linux 7 (Core). The system is equipped with two Intel(R) Xeon(R) Gold 6230R CPUs and 384GB of total RAM. The experiment used four NVIDIA A100 GPUs (40 GB each). The software stack includes Python 3.10.16, PyTorch 2.6.0, and CUDA 11.8. This setup was used for all model training and evaluation.

\subsection{Training Setup}

All models are trained using the AdamW optimizer with a learning rate of $10^{-3}$ and weight decay of $10^{-5}$. We employ a OneCycle learning rate scheduler over $500$ training epochs. The batch size is set to $32$. For benchmark experiments and the \emph{different data size} experiment, we use the resolution of $128 \times 128$ and time step size of $0.2$. For the \emph{different observation time length} experiment and the \emph{different time step size} experiment, we use the resolution of $64 \times 64$. For certain initializations that led to gradient explosions, we applied gradient clipping during training.

\section{Additional Results}

\subsection{Visualization of predicted solutions}
\label{section:visualization}
\begin{figure}[htbp]
    \centering
    \includegraphics[width=\textwidth]{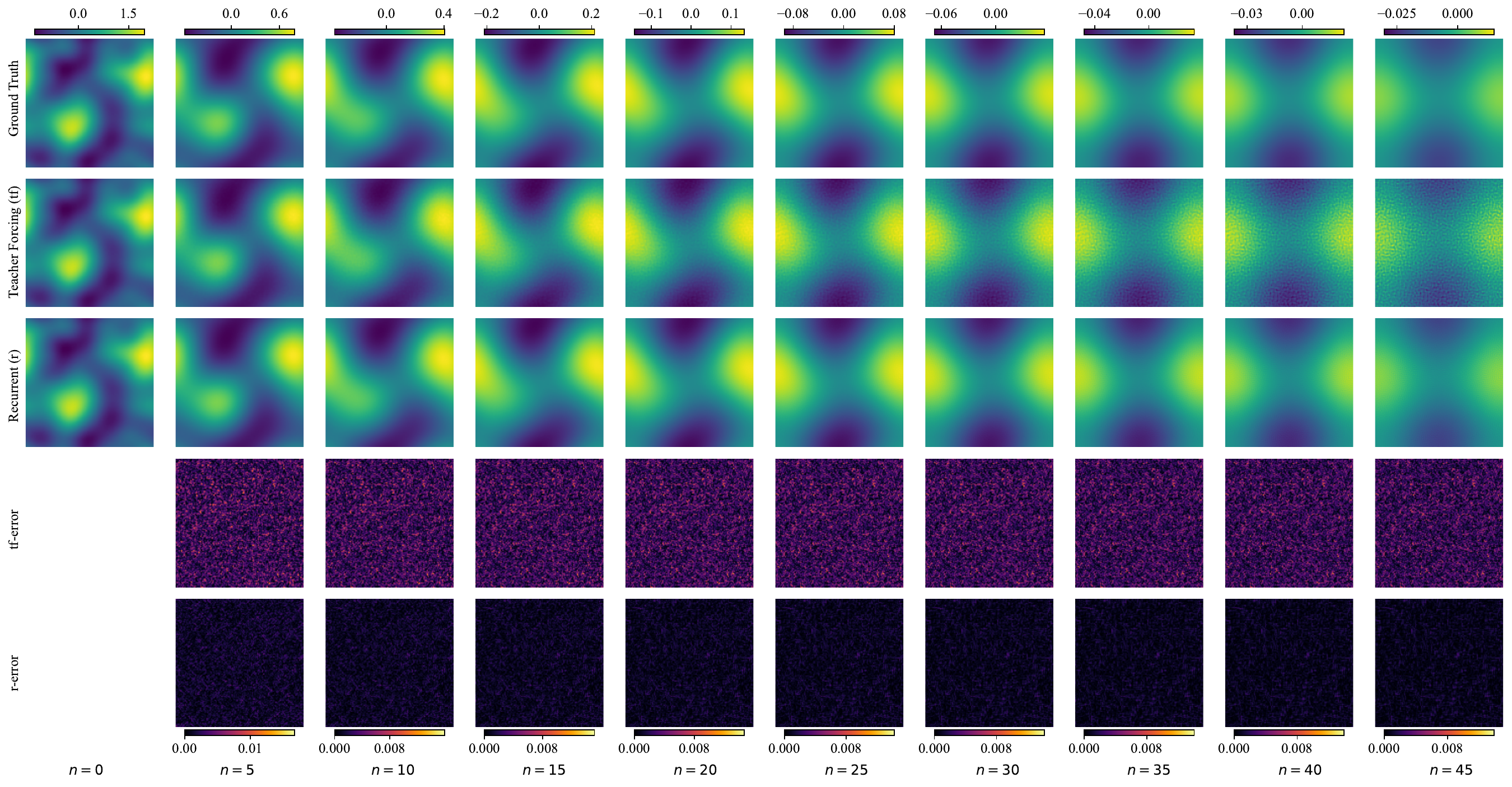}
    \caption{Visualization of predicted solutions for the transient heat conduction problem. The base model is MgNO.}
\end{figure}

\begin{figure}[htbp]
    \centering
    \includegraphics[width=\textwidth]{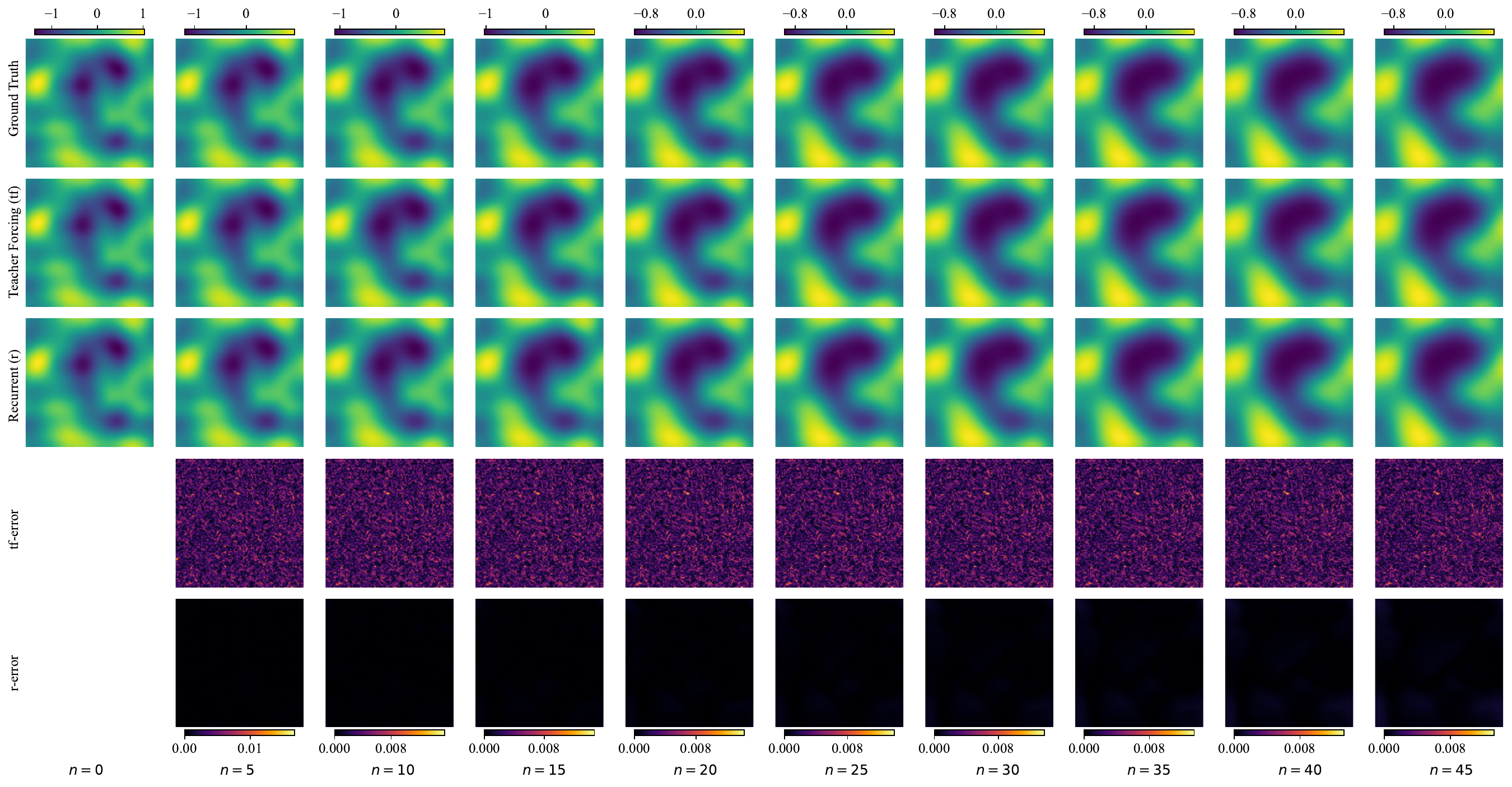}
    \caption{Visualization of predicted solutions for the Allen-Cahn problem. The base model is MgNO.}
\end{figure}

\begin{figure}[htbp]
    \centering
    \includegraphics[width=\textwidth]{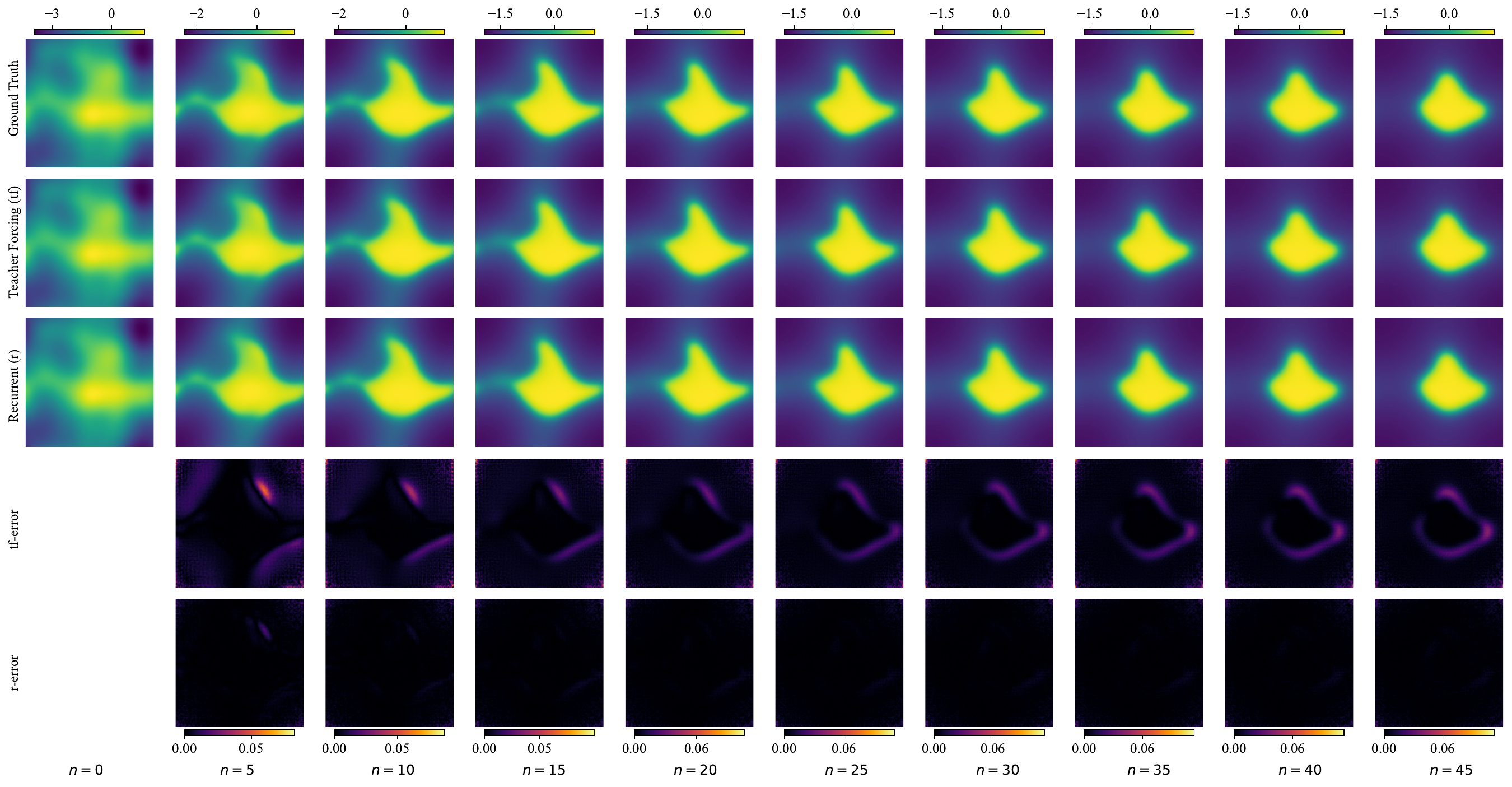}
    \caption{Visualization of predicted solutions for the Cahn-Hilliard problem. The base model is MgNO.}
\end{figure}

\begin{figure}[htbp]
    \centering
    \includegraphics[width=\textwidth]{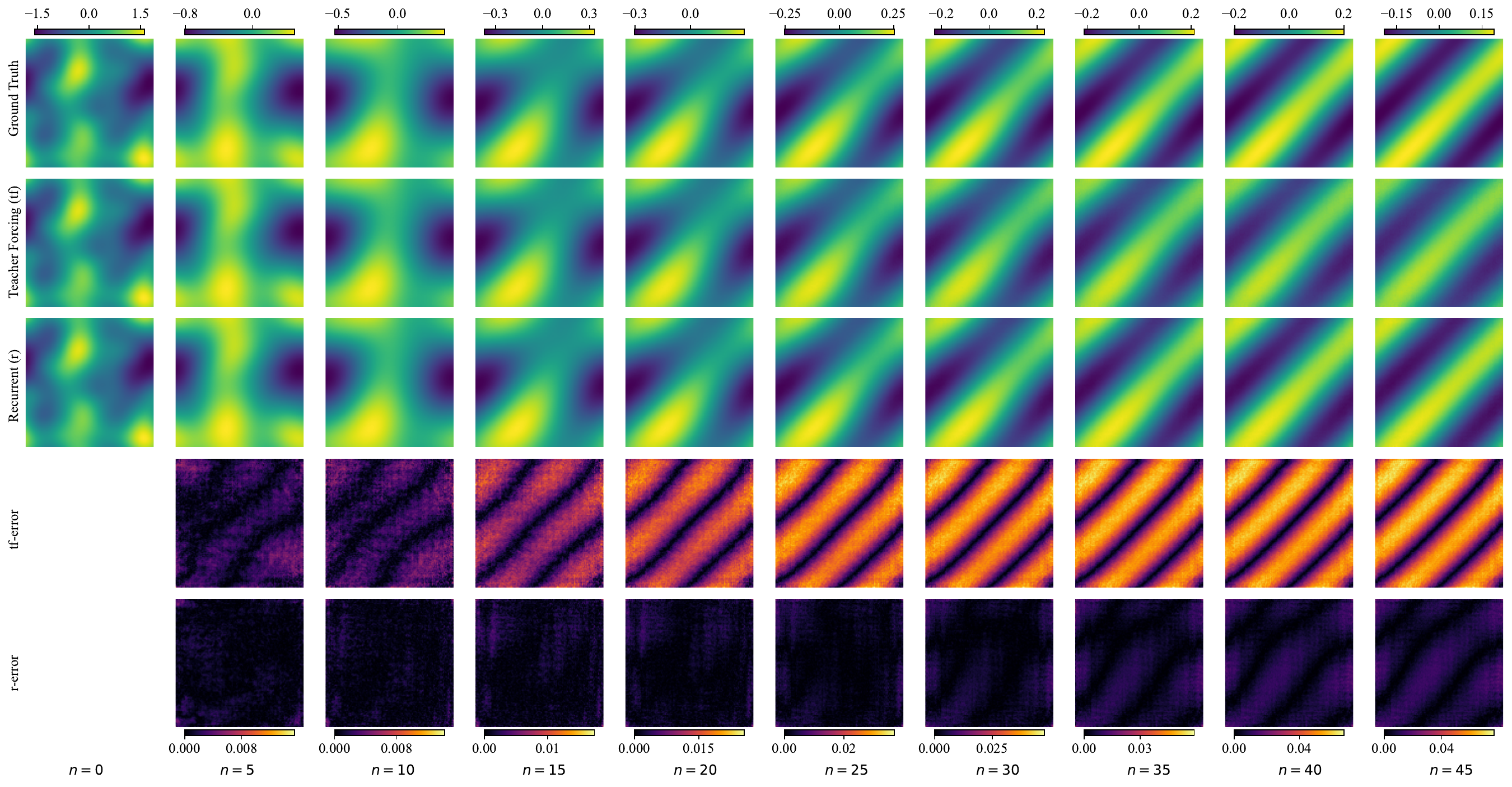}
    \caption{Visualization of predicted solutions for the Navier-Stokes problem. The base model is MgNO.}
\end{figure} 
\end{document}